\documentclass[12pt]{article}

\usepackage{graphicx}
\usepackage{subfigure}
\usepackage{booktabs} 
\usepackage{hyperref}

\usepackage[T1]{fontenc}    
\usepackage{url}            
\usepackage{amsfonts}       
\usepackage{nicefrac}       
\usepackage[protrusion=true,expansion=true]{microtype}

\usepackage[latin9]{inputenc}
\usepackage{amsmath}
\usepackage{amssymb}
\usepackage{algorithm}
\usepackage[noend]{algpseudocode}
\usepackage[disable]{todonotes}

\usepackage{natbib}
\usepackage{authblk}

\newtheorem{theorem}{Theorem}

\newtheorem{proposition}[theorem]{Proposition}

\newcommand{\beq}{\begin{equation}}
\newcommand{\eeq}{\end{equation}}

\let\hat\widehat

\begin{document}

\let\hat\widehat

\title{Cautious Deep Learning}

\author[1]{Yotam Hechtlinger\thanks{yhechtli@stat.cmu.edu}}
\author[2]{Barnab\'as P\'oczos\thanks{bapoczos@cs.cmu.edu }}
\author[1]{Larry Wasserman\thanks{larry@cmu.edu}}
\affil[1]{Department of Statistics and Data Science, Carnegie Mellon University}
\affil[2]{Machine Learning Department, Carnegie Mellon University}

\renewcommand\Authands{ and }

\maketitle

\begin{abstract}
Most classifiers operate by selecting the maximum of an estimate of the conditional distribution 
$p(y|x)$ where $x$ stands for the features of the instance to be classified and $y$ denotes its label. This often results in a {\em hubristic bias}: overconfidence in the assignment of a definite label.
Usually, the observations are concentrated on a small volume 
but the classifier provides definite predictions for the entire
space. We propose
constructing conformal prediction sets \citep{vovk2005algorithmic} which contain a set of labels rather than a single label.
These conformal prediction sets contain the true label with probability $1-\alpha$.
Our construction is based on
$p(x|y)$ rather than $p(y|x)$
which results in a classifier that
is very cautious: it outputs the null set ---
meaning ``I don't know'' ---
when the object does not resemble the training examples.
An important property of our approach is that adversarial attacks
are likely to be predicted as the null set or would also include the true label.
We demonstrate
the performance on the ImageNet ILSVRC dataset and the CelebA and IMDB-Wiki facial datasets using high dimensional
features obtained from state of the art convolutional neural networks. 
\end{abstract}

\section{Introduction}

We consider multiclass classification 
with a feature space $\mathcal{X}$ 
and labels
$\mathcal{Y}=\left\{ 1,\ldots,k\right\}.$ 
Given the training data
$(X_1,Y_1),\ldots, (X_n,Y_n)$,
the usual goal is to find a prediction function
$\hat F:\mathbb{\mathcal{X}\longmapsto\mathcal{Y}}$ 
with low classification error
$P(Y\neq \hat F(X))$ where $(X,Y)$ is a new observation of an input-output pair.
This type of prediction produces a definite prediction even
for cases that are hard to classify.

In this paper we use conformal prediction \citep{vovk2005algorithmic}
where we estimate a set-valued function
$C:\mathcal{X} \longmapsto2^{\mathcal{Y}}$ 
with the guarantee that
$P(Y\in C\left(X\right)) \geq1-\alpha$ for all distributions $P$. 
This is a distribution-free confidence guarantee.
Here, $1-\alpha$ is a user-specified confidence level.
We note that
the ``classify with a reject option'' \citep{herbei2006classification} 
also allows set-valued predictions but does not give 
a confidence guarantee.

The function $C$ can sometimes output
the null set. That is,
$C(x) = \emptyset$ for some values of $x$.
This allows us to distinguish two types of uncertainty.
When $C(x)$ is a large set, there are many possible labels consistent with $x$.
But when $x$ does not resemble the training data,
we will get $C(x) = \emptyset$ alerting us that we have not seen examples
like this so far.

There are many ways to construct
conformal prediction sets.
Our construction is based on finding an estimate
$\hat p(x|y)$ of $p(x|y)$.
We then find an appropriate scalar $\hat t_y$
and we set
$C(x) = \{y:\ \hat p(x|y) > \hat t_y\}$.
The scalars are chosen so that $P(Y\in C(X))\geq 1-\alpha$.
We shall see that this construction works
well when there is a large number of classes
as is often the case in deep learning classification problems.
This guarantees that
$x$'s with low probability --- that is regions where we have not seen training data ---
get classified as $\emptyset$.

An important property of this approach is that 
$\hat p(x|y)$
can be estimated independently for each class. Therefore, $x$ is
predicted to a given class in a standalone fashion which enables
adding or removing classes without the need to retrain the
whole classifier. In addition, we empirically demonstrate that the
method we propose is applicable to large-scale high-dimensional data by applying
it to the ImageNet ILSVRC dataset and the CelebA and IMDB-Wiki facial datasets using features obtained from state
of the art convolutional neural networks.

{\bf Paper Outline.}
In section \ref{sec:P_X_given_Y} we discuss the difference between
$p(y|x)$ and $p(x|y)$. In section~\ref{sec:Motivation} we provide an example to enlighten our
motivation. In section~\ref{sec:Conformal_Prediction} we present the
general framework of conformal prediction and survey relevant works in
the field. In section \ref{sec:Method} we formally present our
method. In section \ref{sec:Examples} we demonstrate the performance
of the proposed classifier on the ImageNet challenge dataset using state of the
convolutional neural networks. In section \ref{sec:gender_recognition} we consider the problem of gender classification from facial pictures and show that even when current classifiers fail to generalize from CelebA dataset to IMDB-Wiki dataset, the proposed classifier still provides sensible results. Section \ref{sec:Discussion} contains our
discussion and concluding remarks.
The Appendix in the supplementary material contains some technical details.

{\bf Related Work.}
There is an enormous literature
on set-valued prediction.
Here we only mention some of the most relevant references.
The idea of conformal prediction originates from
\citet{vovk2005algorithmic}.
There is a large followup literature due to Vovk and his colleagues
which we highly recommend for the interested readers.
Statistical theory for conformal methods
was developed in
\citep{lei2014classification, lei2013distribution, lei2014distribution}, and
the multiclass case was studied in
\citep{sadinle2017least}
where the goal was to develop small prediction sets
based on estimating $p(y|x)$.
The authors of that paper, similarly to \citet{vovk2003mondrian}, tried to avoid outputting
null sets. In this paper, we use this as a feature.
Finally, we mention a related but different technique
called
classification with the ``reject option'' \citep{herbei2006classification}.
This approach permits one to sometimes refrain from providing a classification but it does
not aim to give confidence guarantees. 

Recently, \citet{lee2018simple} suggested a framework based on $p(x|y)$ to predict out of distribution and adversarial attacks.

\section{$p(y|x)$ Versus $p(x|y)$}
\label{sec:P_X_given_Y}

Most classifiers --- including most conformal classifiers ---
are built by estimating $p(y|x)$.
Typically one sets the predicted label of a new $x$ to be
$\hat{f}\left(x\right)=\arg\max_{y\in\mathcal{Y}}\left\{\hat p(y|x)\right\}$.
Since
$p(y|x) = p(x|y)p(y)/p(x)$
the prediction
involves the balance between
$p(y)$ and $p(x|y)$.
Of course, in the special case
$p(y)=1/k$ for all $y$,
we have
$\arg\max_{y\in\mathcal{Y}}\{ p(y|x)\} = \arg\max_{y\in\mathcal{Y}}\{ p(x|y)\}$.

However, for set-valued classification, $p(y|x)$ can be negatively affected
by $p(y)$ and $p(x|y)$.  Indeed, in this case there are significant advantages to using
$p(x|y)$ to construct the classifier.  Taking $p(y)$ into account ties
the prediction of an observation $x$ with the likelihood of observing
that class. Since there is no restriction on the number of
classes, ultimately an observation should be predicted to a class
regardless of the class popularity.  Normalizing by $p(x)$
makes the classifier oblivious to the probability of actually
observing $x$. When $p(x)$ is extremely low (an outlier), 
$p(y|x)$ still selects the most likely label
out of all tail events. In practice this may result with most of the
space classified with high probability to a handful of classes almost
arbitrarily despite the fact that the classifier has been presented
with virtually no information in those areas of the space. This
approach might be necessary if a single class has to be selected
$\forall x\in\mathcal{X}$. However, if this is not the case, then a reasonable prediction for an $x$ with small $p(x)$ is the null set. 

There are also conformal methods utilizing $p(y|x)$ to predict a set of classes \citep{sadinle2017least, vovk2003mondrian}. There methods do not overcome the inherent weakness within $p(y|x)$. As will be explained later on, the essence of this methods is to classify $x$ to $C\left(x\right)=\left\{ y\mid P\left(y\mid x\right)\geq t\right\}$ for some threshold t. Due to the nature of $p(y|x)$ the points which are most likely to be predicted as the null set are when  $P(y=j|x)=\frac{1}{k}$, for all classes $j\in\mathcal{Y}$. But this is exactly the points in space for which any set valued prediction should predict all class as possible.  

As we shall see, conformal predictors based on $p(x|y)$ can overcome all these issues.

\section{Motivating Example - Iris Dataset}\label{sec:Motivation}

The Iris flower data set is a benchmark dataset often used to demonstrate
classification methods. It contains four features that were measured
from three different Iris species. In this example, for visualization
purposes, we only use two features: the sepal and petal lengths in cm.

\begin{figure*}
\begin{tabular}{ccc}
\includegraphics[width=0.31\textwidth]{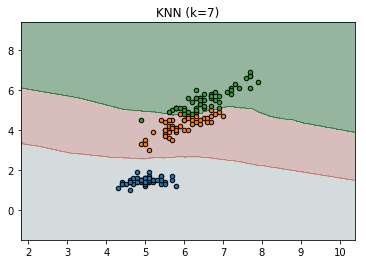} &   
\includegraphics[width=0.31\textwidth]{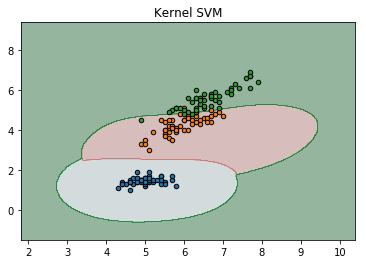} &   
\includegraphics[width=0.31\textwidth]{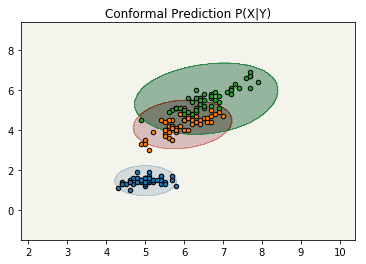} \\
(a) & (b) & (c)\\[6pt]
\end{tabular}
\caption{Classification boundaries for different methods for the Iris
dataset. For the conformal prediction method (c) (with $\alpha=0.05$) the
overlapping areas are classified as multiple classes and white areas
are classified as the null set. For the standard methods (a-b), the decision
boundaries can change significantly with small changes in some of the data points and the prediction cannot be justified in most
of the space. Online version in color.}
\label{fig:Iris_results}
\end{figure*}

Figure \ref{fig:Iris_results} shows the decision boundaries for
this problem comparing the results of (a) K-nearest neighbors (KNN),
(b) support vector machines with the RBF kernel (SVM) and (c) our
conformal prediction method using an estimate 
$\hat p(x|y)$.

Both the KNN and the SVM methods provide sensible boundaries between
the class where there are observations. In areas with low density
$p(x)$ the decision boundaries are significantly
different. The SVM classifies almost all of the space to a single
class. The KNN creates an infinite strip bounded between two (almost
affine) half spaces. In a hubristic manner, both methods provide very
different predictions with probability near one without sound
justification.

The third plot shows the conformal set
$C(x) = \{y:\ \hat p(x|y) > \hat t_y\}$
where
the $\hat t_y$ is chosen 
as described in Section~\ref{sec:Method}.
The result is a cautious prediction. 
If a new $X$ falls into a region with little training data
then we output $\emptyset$.
In such cases our proposed method modestly avoids providing any claim.

\section{Conformal Prediction}\label{sec:Conformal_Prediction}

Let $(X_1,Y_1),\ldots, (X_n,Y_n)$ be $n$ independent and identically distributed (iid) pairs of observations
from a distribution $P$. In set-valued supervised prediction, the
goal is to find a set-valued function $C(x)$ such that
\beq\label{eq::required}
P(Y\in C(X)) \ge 1-\alpha,
\eeq
where $(X,Y)$ denotes a new pair of observations.

Conformal prediction --- a method created by Vovk and collaborators
\citep{vovk2005algorithmic} --- provides a general approach to
construct prediction sets based on the observed data without any
distributional assumptions. 
The main idea is to construct a 
{\rm conformal score},
which is a real-valued, permutation-invariant function
$\psi(z, {\cal D})$
where $z=(x,y)$ and
${\cal D}$ denotes the training data.
Next we form an augmented dataset
${\cal D}'= \{(X_1,Y_1),\ldots, (X_n,Y_n),(X_{n+1},Y_{n+1})\}$
where
$(X_{n+1},Y_{n+1})$ is set equal to arbitrary values $(x,y)$.
We then define
$R_i = \psi( (X_i,Y_i), {\cal D}')$
for $i=1,\ldots, n+1$.
We test the hypothesis $H_0: Y=y$ that the new label $Y$ is equal to $y$
using the p-value
$\pi(x,y) = 1/(n+1)\sum_{i=1}^{n+1} I( R_i \geq R_{n+1})$.
Then we set
$C(x) = \{y:\ \pi(x,y)\geq \alpha\}$.
\citep{vovk2005algorithmic} proves that
$P(Y\in C(X))\geq 1-\alpha$ for all distributions $P$.
There is a great flexibility in the choice
of conformity score and \ref{conformal_examples} discusses important examples. 

As described above,
it is computationally expensive to construct
$C(x)$ since we must re-compute
the entire set of conformal scores for each choice of $(x,y)$.
This is especially a problem
in deep learning applications where training is usually expensive.
One possibility for overcoming the computational burden is based
on data splitting where
$\hat p(x|y)$ is estimated from part of the data and the conformal scores
are estimated from the remaining data; see 
\citep{vovk2015cross,lei2014distribution}. 
Another approach
is to construct the scores from the original data
without augmentation.
In this case, we no longer have
the finite sample guarantee $P(Y\in C(X))\geq 1-\alpha$ for all distributions $P$,
but we do get
the asymptotic 
guarantee $P(Y\in C(X))\geq 1-\alpha- o_P(1)$ 
as long as some conditions are satisfied\footnote{A sequence of random variables $X_{1},X_{2},\ldots,$ is $o_{p}\left(1\right)$ if $\forall\epsilon>0,$ $\lim_{n\rightarrow\infty}P\left(\left|X_{n}\right|\geq\epsilon\right)=0$.}. See \citep{sadinle2017least} for further discussion on
this point.

\subsection{Examples}\label{conformal_examples}

Here are several known examples for conformal methods used on different problems.

{\bf Supervised Regression.}  Suppose we are interested in the
supervised regression problem. Let
$\hat{f}:\mathcal{X}\rightarrow\mathcal{Y}$ be any regression function
learned from training data. Let $\epsilon_i$ denote the residual error of
$\hat{f}$ on the observation $i$, that is,
$\epsilon_i =|\hat{f}\left(X_{i}\right)-Y_{i}|$.
Now we form the ordered residuals
$\epsilon_{(1)} \leq \cdots \leq \epsilon_{(n)}$, and then define
$$ 
C(x)=\left\{y:\ 
|\hat{f}\left(x\right)-y| \leq \epsilon_{\left(\lceil\left(1-\alpha\right)\cdot   n\rceil\right)}\right\}.
$$
If $\hat f$ is a consistent estimator of $\mathbb{E}[Y|X=x]$ then
$P(Y\in C(X)) = 1-\alpha + o_P(1)$. See \citet{lei2014distribution}.

{\bf Unsupervised Prediction.}  Suppose we observe independent and
identically distributed $Y_{i},\ldots,Y_{n}\in\mathbb{R}^{d}$ from
distribution $P$. The goal is to construct a prediction set $C$ 
for new $Y$.
\citet{lei2013distribution} 
use the level set
$C = \{y:\ \hat p(y) > t\}$
where $\hat p$ is a kernel density estimator.
They show that if $t$ is chosen carefully then
$P(Y\in C)\geq 1-\alpha$ for all $P$.

{\bf Multiclass Classification.}  There are two notable solutions also
using conformal prediction for the multiclass classification problem
which are directly relevant to this work.

{\em Least Ambiguous Set-Valued Classifiers with Bounded Error Levels}.
\citet{sadinle2017least} extended the results of \citet{lei2014classification} and defined
$R_i =\hat{p}\left(Y_i\mid X_i\right)$, where $\hat{p}$ is
any consistent estimator of $p(y|x)$.
They defined the \textit{minimal ambiguity} as
$\mathbb{A}\left(C\right)=\mathbb{E}\left(\left|C\left(X\right)\right|\right)$
which is the expected size
of the prediction set.
They proved that out of all the classifiers achieving the
desired $1-\alpha$ coverage, this solution minimizes the
ambiguity.
In addition, the paper considers class specific coverage controlling for every class
$P\left(Y\in C\left(X\right)\mid Y=y\right)\geq1-\alpha_{y}$.

{\em Universal Predictor.}  \citet{vovk2003mondrian}
introduce the concept of universal predictor and provide an explicit way to construct one. A universal predictor is the classifier that produces, asymptotically, no more multiple prediction than any other classifier achieving $1-\alpha$ level coverage. In addition, within the family of all $1-\alpha$ classifiers that produce the minimal number of multiple predictions it also asymptotically obtains at least as many null predictions.

\section{The Method}
\label{sec:Method}

\subsection{The Classifier}

Let $\hat p(x|y)$ be an estimate
of the density $p(x|y)$ for class $Y=y$.
Define $\hat t_y$ to be the empirical
$1-\alpha$ quantile of the values
$\{\hat p(X_i|y)\}$.
That is, 
\begin{equation}
\hat t_y = 
\sup\Biggl\{ y:\ \frac{1}{n_y}\sum_i I( \hat p(X_i|y) \geq t) \geq 1-\alpha \Biggr\}
\end{equation}
where
$n_y = \sum_i I(Y_i = y)$.
Assuming that $n_y \to \infty$
and minimal conditions on
$p(x|y)$ and $\hat p(x|y)$,
it can be shown that
$\hat t_y \stackrel{P}{\to} t_y$ where
$t_y$ is the largest $t$ such that
$\int_{y>t} p(x|y) dx \geq 1-\alpha$.
See \cite{cadre2009clustering} and
\cite{lei2013distribution}.
We set
$C(x) = \{y:\ \hat p(x|y) \geq \hat t_y\}$.
We then have the following proposition which is proved in the appendix.

\begin{proposition}
Assume the conditions in
\cite{cadre2009clustering} stated also in the appendix. Let $(X,Y)$ be a new observation.
Then
$|P(Y\in C(X)) - (1-\alpha)| \stackrel{P}{\to} 0$
as $\min_y n_y \to \infty$.
\end{proposition}

An exact, finite sample method can be obtained using data splitting.
We split the training data into two parts.
Construct $\hat p(x|y)$ from the first part of the data.
Now evaluate $\{\hat p(X_i|y)\}$
on the second part of the data and define $\hat t_y$
using these values.
We then set $C(x) = \{y:\ \hat p(x|y) \geq \hat t_y\}$.
We then have:

\begin{proposition}
Let $(X,Y)$ be a new observation.
Then, for every distribution and every sample size,
$P(Y\in C(X))\geq 1-\alpha$.
\end{proposition}

This follows from the theory in
\cite{lei2014distribution}.
The advantage of the splitting approach
is that there are no conditions
on the distribution, and the confidence guarantee
is finite sample. There is no large sample approximation.
The disadvantage is that the data splitting can lead to larger prediction sets.
Algorithm
\ref{alg:Training_Algorithm} describes the training, and Algorithm
\ref{alg:Prediction_Algorithm} describes the prediction.

\begin{algorithm}[]
\caption{Training Algorithm}\label{alg:Training_Algorithm}
\begin{algorithmic}[0]
\State \textbf{Input:} Training data $Z=\left(X,Y\right)$, Class list $\mathcal{Y}$, Confidence level $\alpha$, Ratio $p$.
\State $\hat{p}_{list}=list;\hat{t}_{list}=list$ \Comment{Initialize lists}
\For{$y$ in $\mathcal{Y}$}\Comment{Loop over all the classes independently}
	\State $X_{tr}^{y},X_{val}^{y}\leftarrow SubsetData\left(Z,\mathcal{Y},p\right)$ \Comment{Split $X\mid y$ with ratio $p$}
    \State $\hat{p}_{y}\leftarrow LearnDensityEstimator\left(X_{tr}^{y}\right)$ 
    \State $\hat{t}_{y}\leftarrow Quantile\left(\hat{p}_{y}\left(X_{val}^{y}\right),\alpha\right)$ \Comment{The validation set $\alpha$ quantile}
    \State $\hat{p}_{list}.append\left(\hat{p}_{y}\right);\hat{t}_{list}.append\left(\hat{t}_{y}\right)$
\EndFor
\State \textbf{return} $\hat{p}_{list};\ \hat{t}_{list}$
\end{algorithmic}
\end{algorithm}

\begin{algorithm}[]
\caption{Prediction Algorithm}\label{alg:Prediction_Algorithm}
\begin{algorithmic}[0]
\State \textbf{Input:} Input to be predicted $x$, Trained $\hat{p}_{list};\ \hat{t}_{list}$, Class list $\mathcal{Y}$.
\State $C=list$ \Comment{Initialize $C\left(x\right)$}
\For{$y$ in $\mathcal{Y}$}\Comment{Loop over all the classes independently}
	\If{$\hat{p}_{y}\left(x\right)\geq\hat{t}_{y}$}
    	\State $C.append\left(y\right)$
    \EndIf
\EndFor
\State \textbf{return} $C$
\end{algorithmic}
\end{algorithm}

\subsection{Density Estimation}\label{subsec:density_estimation}

The density $p(x|y)$
has to be estimated from data. One possible way is to use the standard kernel density
estimation method, which was shown to be optimal in the conformal setting 
under weak conditions in \citet*{lei2013distribution}. This is useful for theoretical purposes 
due to the large literature on the topic. Empirically, it is faster to use the 
distance from the $k$ nearest neighbors. 

Density estimation in high dimensions is a difficult problem.
Nonetheless, as we will show in the numerical experiments (Section~\ref{sec:Examples}),
 the proposed method works well in these tasks as well. An intuitive reason for this could be that the accuracy of the conformal prediction
does not actually require $\hat p(x|y)$ to be close to
$p(x|y)$ in $L_2$.
Rather, all we need is that
the ordering imposed by
$\hat p(x|y)$ approximates the ordering defined by
$p(x|y)$.
Specifically,
we only need that
$\{(x,x'):\ p(x|y) > p(x'|y)+ \Delta \}$
is approximated by
$\{(x,x'):\ \hat p(x|y) > \hat p(x'|y)+ \Delta \}$
for $\Delta>0$.
We call this ``ordering consistency.''
This is much weaker than
the usual requirement that
$\int (\hat p(x|y) - p(x|y))^2 dx$ be small. This new definition and implications on the approximation of $p\left(x\mid y\right)$ will be further expanded in future work. 

\subsection{Class Adaptivity}

As algorithms \ref{alg:Training_Algorithm} and
\ref{alg:Prediction_Algorithm} demonstrate, the training and
prediction of each class is independent from all other classes. This
makes the method adaptive to addition and removal of classes
ad-hoc. Intuitively speaking, if there is $1-\alpha$ probability for
the observation to be generated from the class it will be classified
to the class regardless of any other information.

Another desirable property of the method is that it is possible to
obtain different coverage levels per class if the task requires
that. This is achieved by 
setting $\hat t_y$ to be the
$1-\alpha_y$ quantile
of the values
$\{ \hat p(X_i|y)\}$.

\subsection{Class Interaction}
Defining $p(x|y)$ independently for each class has the desired property of
class adaptivity, but also it discards relevant information regarding the relations of each of the classes. Figure \ref{fig:Iris_results} (c) demonstrate how the different classifiers decision boundaries are independent. 

\begin{figure}
\centering
\includegraphics[width=0.95\textwidth]{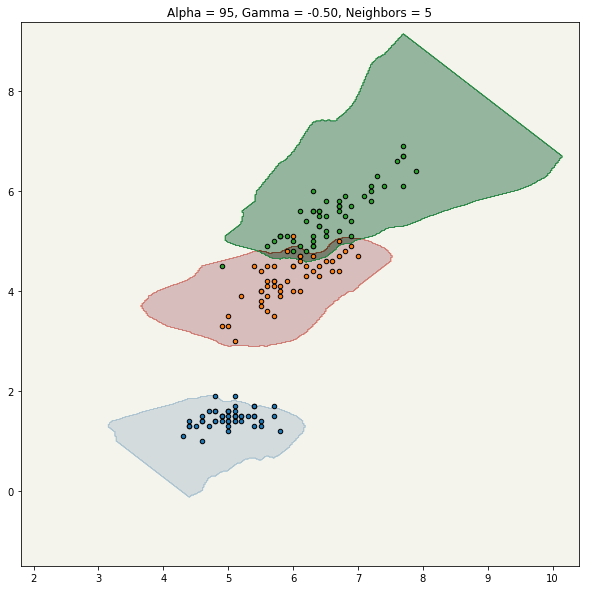} 
\caption{An example of more complex estimation function of $p(x|y)$ providing $0.95$ coverage on the Iris dataset. $p(x|y)$ is higher for points which has high density within the class and low density in the other classes. The decision boundary closely resemble standard methods, while still providing cautious prediction and robustness to out of sample predictions. }
\label{fig:complex_functions}
\end{figure}

More complex decision boundaries can be created using correlated estimators of $p(x|y)$. 
One example of such an estimator is $p'(x|y)\propto p(x|y)-\lambda\sum_{y^{'}\not=y}p(x|y^{'})$, for some $\lambda \in \mathbb{R}$. 
This estimator penalizes high density regions for the other classes. Figure \ref{fig:complex_functions} visualize the results of such estimator on the Iris dataset. 

\section{ImageNet Challenge Example}\label{sec:Examples}

The ImageNet Large Scale Visual Recognition Challenge (ILSVRC)
\citep{deng2009imagenet} is a large visual dataset of more than $1.2$
million images labeled across $1,000$ different classes. It is
considered a large scale complex visual dataset that reflects object
recognition state-of-the-art through a yearly competition.

In this example we apply our conformal image classification method to the ImageNet dataset. We remove
the last layer from the pretrained Xception convolutional neural
network \citep{chollet2016xception} and use it as a feature
extractor. Each image is represented as a $2,048$ dimensional feature
in $\mathbb{R}^{2048}.$ We learn for each of the $1,000$ classes a
unique kernel density estimator trained only on images within the training set of the given class. When we evaluate results of standard methods we
use the Inception-v4 model \citep{szegedy2017inception} to avoid
correlation between the feature extractor and the prediction outcome
as much as possible.

The Xception model obtains near state-of-the-art results of $0.79$
(top-$1$) and $0.945$ (top-$5$) accuracy on ImageNet validation
set. As a sanity check to the performance of our method, selecting for
each image the highest (and top 5) prediction of $\hat{p}\left(x\mid
y\right)$ achieves $0.721$ (top-$1$) and $0.863$ (top-$5$) on ImageNet
validation set. We were pleasantly surprised by this result. Each of the
$\hat{p}\left(x\mid y\right)$'s were learned independently
possibly discarding relevant information on the relation between the classes. The kernel density
estimation is done in $\mathbb{R}^{2,048}$ and the default bandwidth
levels were used to avoid overfitting the training set. Yet the naive
performance is roughly on par with GoogLeNet \citep{szegedy2015going}
the winners of $2014$ challenge (top-1: $0.687$, top-5: $0.889$).

\begin{figure*}
\centering
\begin{tabular}{ccc}
  \includegraphics[width=0.31\textwidth]{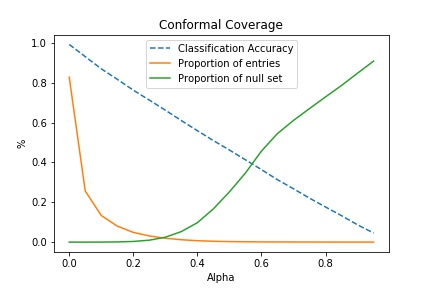} &   \includegraphics[width=0.31\textwidth]{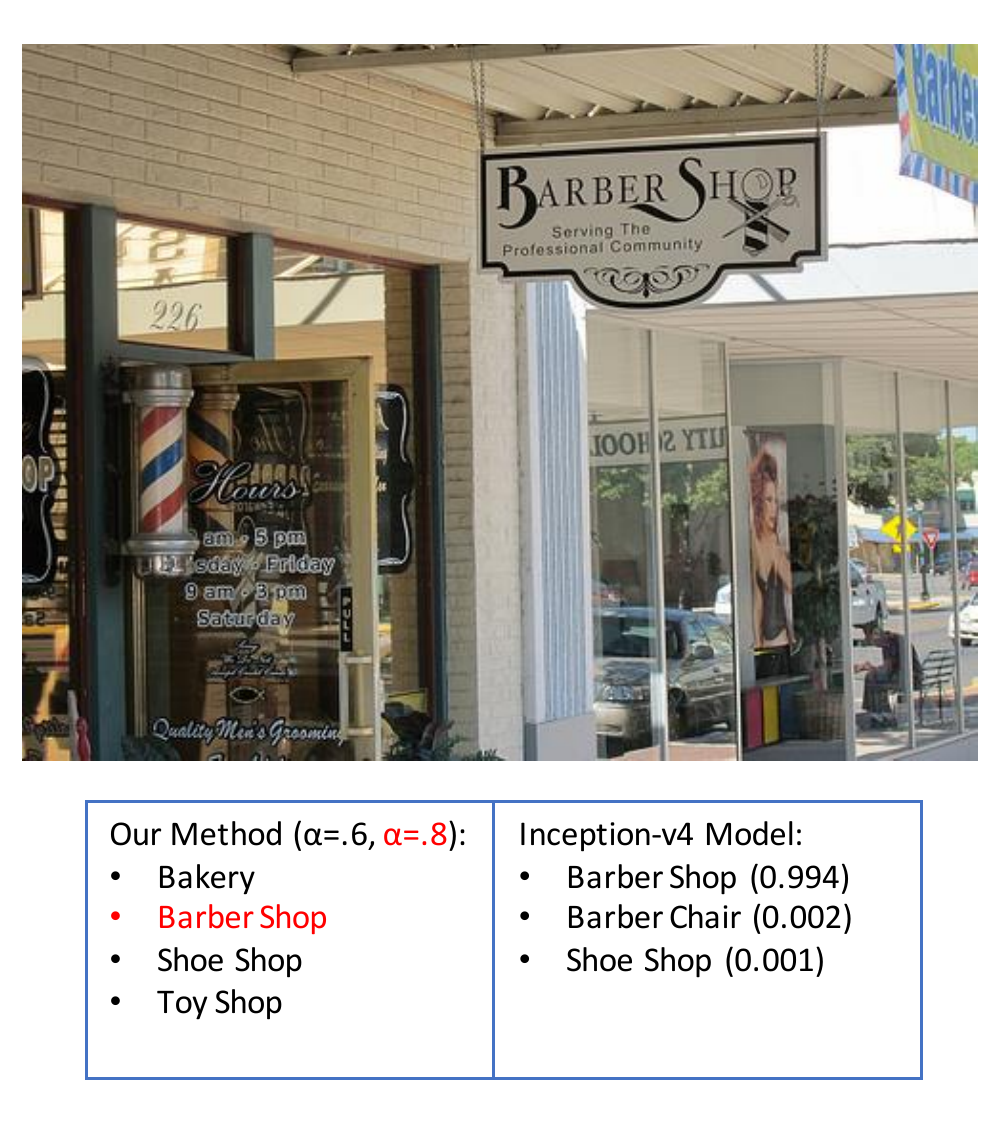} & 
  \includegraphics[width=0.31\textwidth]{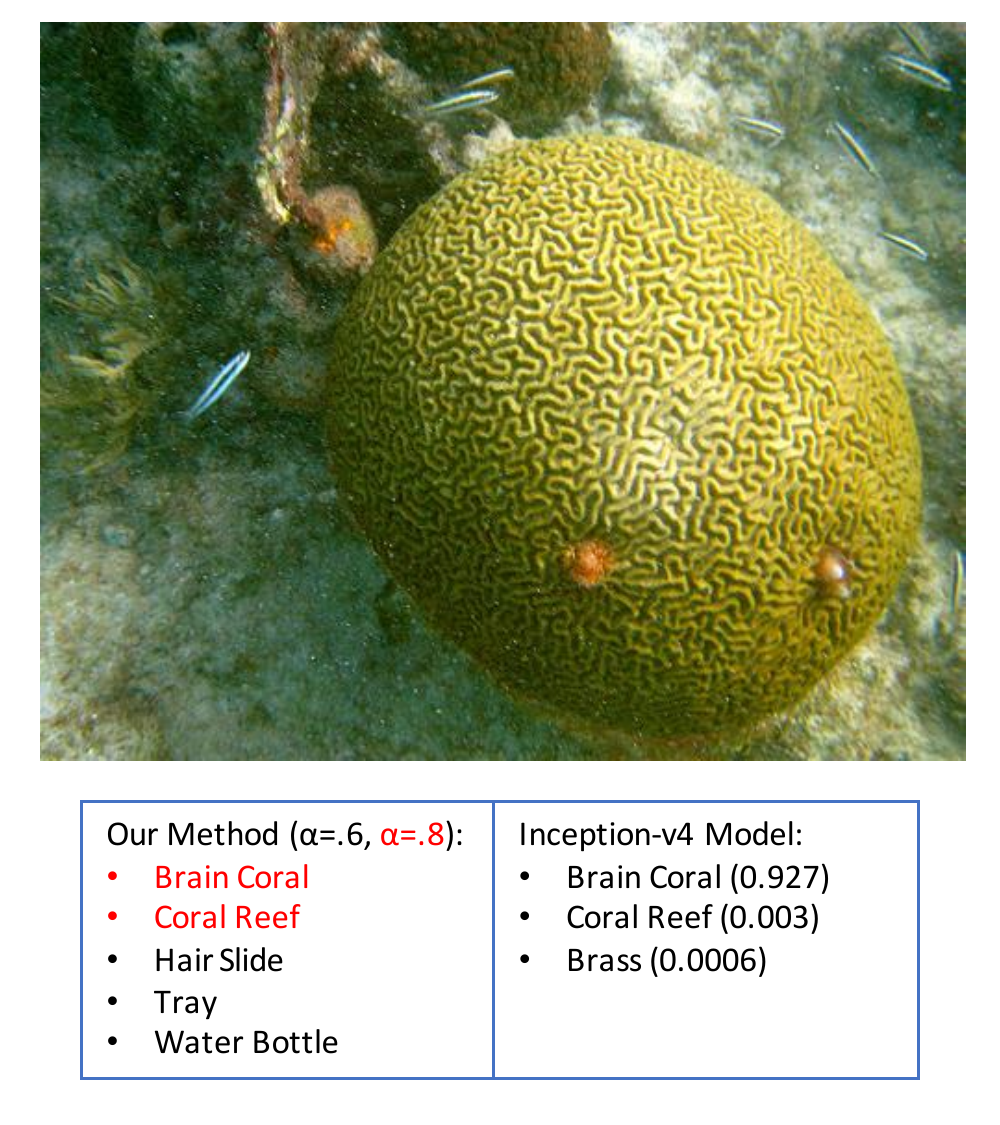} \\
(a) & (b) & (c)\\[6pt]
\end{tabular}
\caption{(a) Performance plot for the conformal method. Accuracy is empirically linear as a function of $\alpha$ but affect the number of classes predicted per sample. (b-c) are illustrative examples. When $\alpha=0.6$ both black and red classes are predicted. When $\alpha=0.8$ the red classes remain.}
\label{fig:Convergence_and_examples}
\end{figure*}

For conformal methods the confidence level is predefined.
The method calibrates the number of classes in the prediction sets
to satisfy the desired accuracy level. The the main component affecting
the results is the hyperparameter $\alpha$. For small values of
$\alpha$ the accuracy will be high but so does the number of classes
predicted for every observation. For large values of $\alpha$ more
observations are predicted as the null set and less observations
predicted per class. Figure \ref{fig:Convergence_and_examples} (a)
presents the trade-off between the $\alpha$ level and the number
of classes and the proportion of null set predictions for this example.
For example $0.5$, accuracy would require on average $2.7$ predictions
per observation and $0.252$ null set predictions. The actual selection
of the proper $\alpha$ value is highly dependent on the task. As
discussed earlier, a separate $\alpha_y$ for each class
can also be used to obtain different accuracy per
class.

Figures \ref{fig:Convergence_and_examples} (b) and (c) show
illustrative results from the ImageNet validation set. (b) presents a
picture of a "\textit{Barber Shop}". When $\alpha=0.6$ the method
correctly suggests the right class in addition to several other
relevant outcomes such as "\textit{Bakery}". When $\alpha=0.8$ only
the "\textit{Barber Shop}" remains. (c) show a "\textit{Brain Coral}". For $\alpha=0.6$ the method still suggests classes which are clearly wrong. As $\alpha$ increases the number of classes decrease and for $\alpha=0.8$ only "\textit{Brain Coral}" and
"\textit{Coral Reef}" remains, both which are relevant. At
$\alpha=0.9$ "\textit{Coral Reef}" remains, which represents a
misclassification following from the fact that the class threshold is
lower than that of "\textit{Brain Coral}". Eventually at $\alpha=0.95$
the null set is predicted for this picture.

\begin{figure}
\centering
\vspace*{-2cm}
\includegraphics[width=1\textwidth]{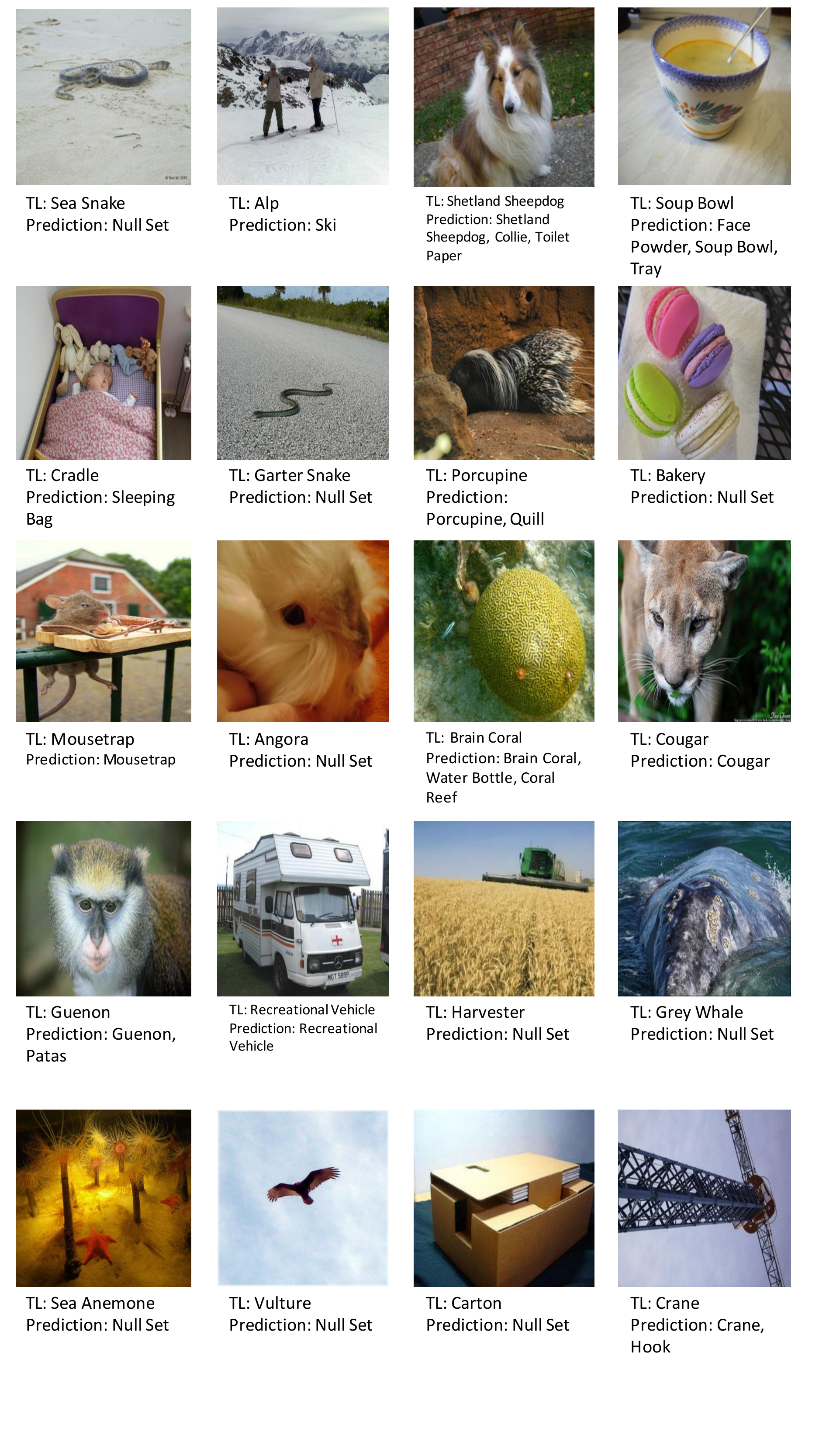}  
\vspace*{-2cm}
\caption{A collage of the first $20$ images in the ImageNet validation set with $\alpha=0.7$. \textit{TL} denotes the image true label and \textit{Prediction} is the method output. By design only $0.3$ accuracy is expected, yet both the true and the false predictions are reasonable. Online version in color.}
\label{fig:Collage}
\end{figure}

Figure \ref{fig:Collage} shows a collage of $20$ images using
$\alpha=0.7$. To avoid selection bias we've selected the first $20$
images in the ImageNet validation set.

\subsection{Adversarial Robustness}\label{sec:adversarial_examples}
Adversarial attacks attempt to fool machine learning models through malicious input. The suggested method is designed to be cautious and provide multiple predictions under uncertainty which results with a robust performance under different attacks. In this section we use the foolbox library \citep{rauber2017foolbox} to generate different attacks on ImageNet validation and test the performance of the method on the ResNet50 model \citep{he2016deep}. We attack the first $200$ images that are accurately classified by the model. 

Table \ref{table:adv_performance} shows the prediction results of two type of attacks, untargeted (using Deepfool \citep{rauber2017foolbox} and the FGSM attack \citep{kurakin2016adversarial}) and targeted (using L-BFGS-B \citep{tabacof2016exploring} and Projected Gradient Descent (PGD) \citep{kurakin2016adversarial}). The untargeted attacks perturb the image the least in order to find any misclassification. This yields predictions of both the true class and the adversarial class. While the attack reduces the performance of the model, the model is more robust than standard methods. Targeted attacks attempt to predict a specific class given apriori (randomly selected). This requires the attack to create larger modifications to the original image, and as a result the model mostly predict the null set both for the true label and the adversarial label. 

\begin{table}
\centering
\begin{tabular}{|l|cc|cc|}
\hline
Accuracy & \multicolumn{2}{c|}{$1-\alpha=0.5$} & \multicolumn{2}{c|}{$1-\alpha=0.75$} \\
& True & Adversarial. & True & Adversarial \\
\hline
DeepFool (Untargeted)& $0.510$ & $0.385$ & $0.770$ & $0.680$ \\
FGSM (Untargeted)& $0.495$ & $0.375$ & $0.750$ & $0.720$ \\
L-BFGS-B (Targeted)  & $0.000$ & $0.015$ & $0.080$ & $0.135$ \\
PGD (Targeted)  & $0.105$ & $0.095$ & $0.430$ & $0.255$ \\
\hline 
\end{tabular}
\caption{Accuracy on $200$ adversarial images. True denotes the accuracy over the original class and Adversarial denotes the accuracy on the adversarial class. Untargeted attacks obtain the expected accuracy on the true label. Targeted attacks have a greater effect on the image and are less likely to be predicted in any class. \label{table:adv_performance}}
\end{table}

\subsection{Outliers}
\begin{figure*}
\begin{tabular}{ccc}
  \includegraphics[width=0.31\textwidth]{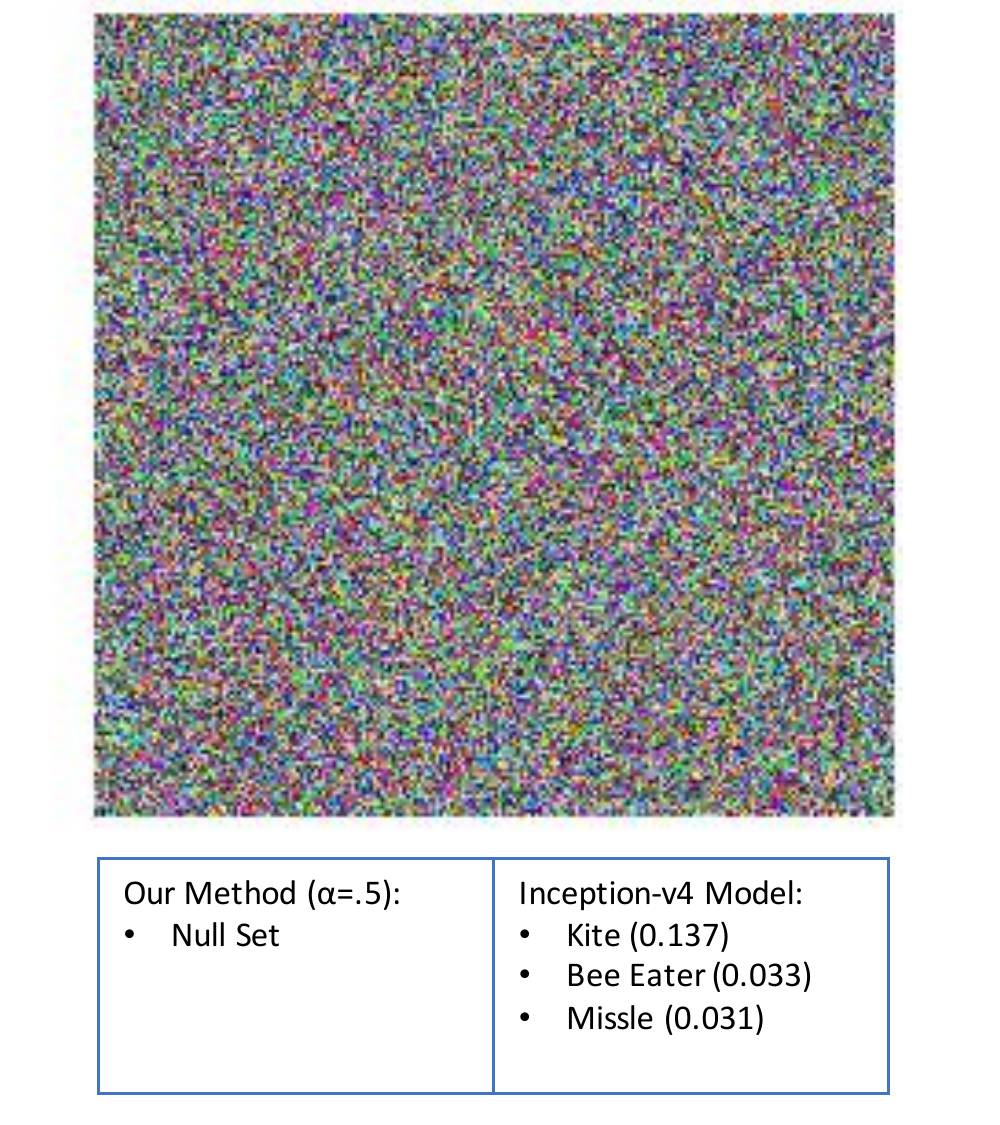} &   \includegraphics[width=0.31\textwidth]{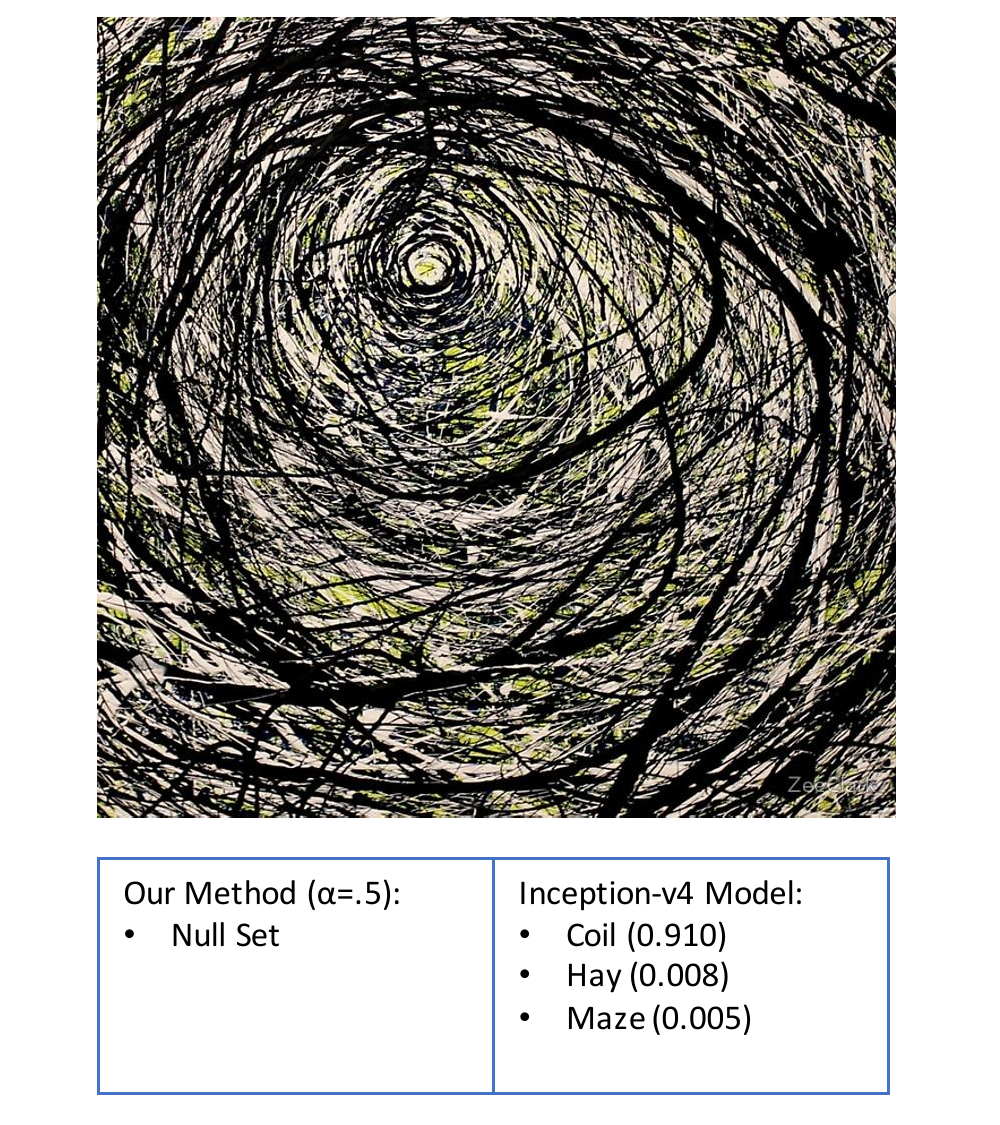} & 
  \includegraphics[width=0.31\textwidth]{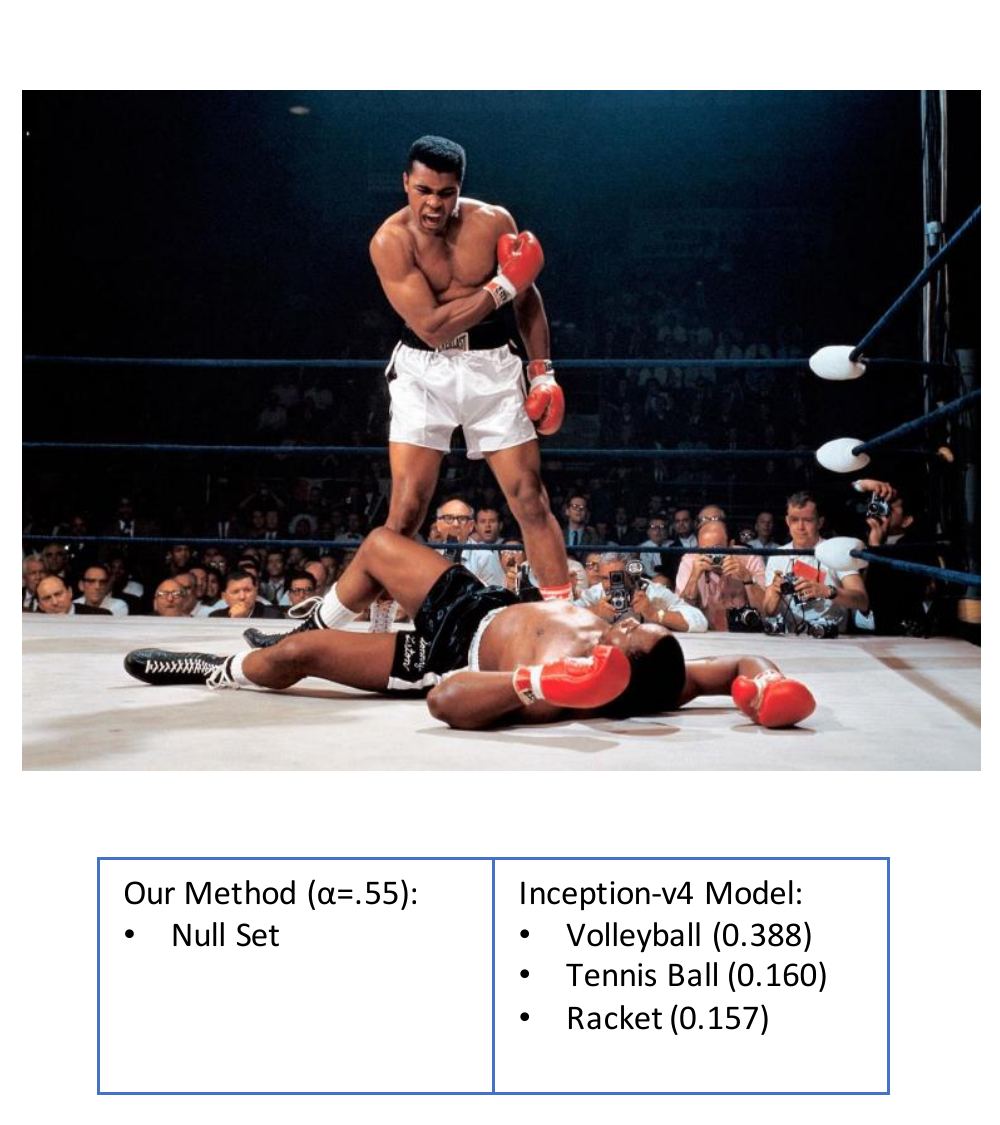} \\
(a) & (b) & (c)\\[6pt]
\end{tabular}
\caption{Classification results for (a) random noise; (b) Jackson Pollock "Rabit Hole"; (c) Muhammad Ali towering over Sonny Liston (1965 rematch). These pictures are outliers for the Imagenet categories. The left labels of each picture are provided by our method and the right are the results of the Inception-v4 model.}
\label{fig:Outliers_Example}
\end{figure*}

Figure \ref{fig:Outliers_Example} (a) shows the outcome when the input is random noise.
We set the threshold $\alpha=0.01$. This
gives a less conservative classifier that should have the
largest amount of false positives. Even with such a low threshold all
$100$ random noise images over $1,000$ categories are correctly
flagged as the null set. Evaluating the same sample on the
Inception-v4 model \citep{szegedy2017inception} results with a top
prediction average of $0.0836$ (with $0.028$ standard error) to "Kite"
and $0.0314$ ($0.009$) to "Envelope". The top-5 classes together has
mean probability of $0.196$, much higher than the uniform distribution
expected for prediction of random noise.

Figure \ref{fig:Outliers_Example} (b) show results on Jackson Pollock
paintings - an abstract yet more structured dataset. Testing $11$
different paintings with $\alpha=0.5$ all result with the null
set. When testing the Inception-v4 model output, $7/11$ paintings are
classified with probability greater than $0.5$ to either "Coil",
"Ant", "Poncho", "Spider Web" and "Rapeseed" depending on the image.

Figure \ref{fig:Outliers_Example} (c) is the famous picture of
Muhammad Ali knocking out Sonny Liston during the first round of
the $1965$ rematch. "Boxing" is not included within in the ImageNet
challenge. Our method correctly chooses the null set with $\alpha$  as
low as $0.55$. Standard method are forced to associate this image with
one of the classes and choose "Volleyball" with $0.38$ probability and
the top-5 are all sport related predictions with $0.781$
probability. This is good result given the constraint of selecting a
single class, but demonstrate the impossibility of trying to create
classes for all topics.

\section{Gender Recognition Example}\label{sec:gender_recognition}

In the next example we study the problem of gender classification from facial pictures. CelebFaces Attributes Dataset (CelebA) \citep{liu2015deep} is a large-scale face attributes dataset with more than $200K$ celebrity images attributed, each with 40 attribute annotations including the gender (Male/Female). IMDB-Wiki dataset is a similar large scale ($500K+$ images) dataset \citep{Rothe-IJCV-2016} with images taken from IMDB and Wikipedia.

\begin{figure*}
\begin{tabular}{ccc}
  \includegraphics[width=0.31\textwidth]{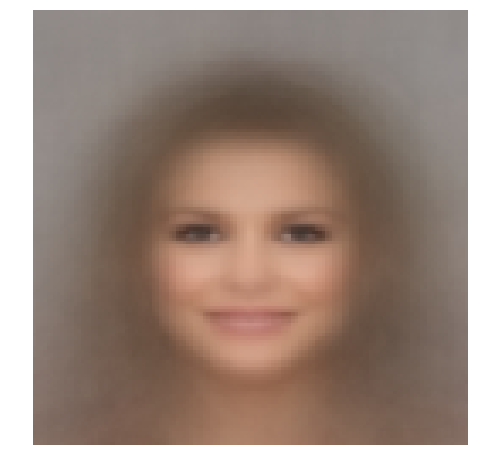} &   \includegraphics[width=0.31\textwidth]{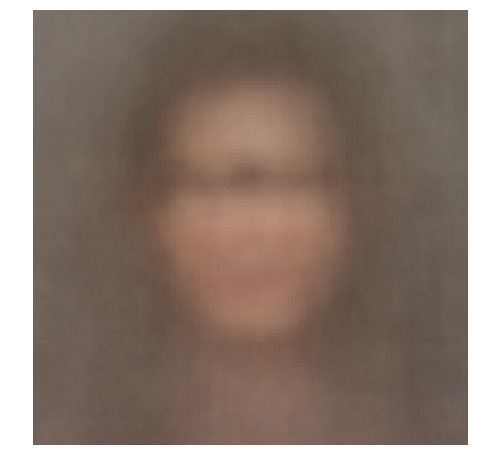} & 
  \includegraphics[width=0.31\textwidth]{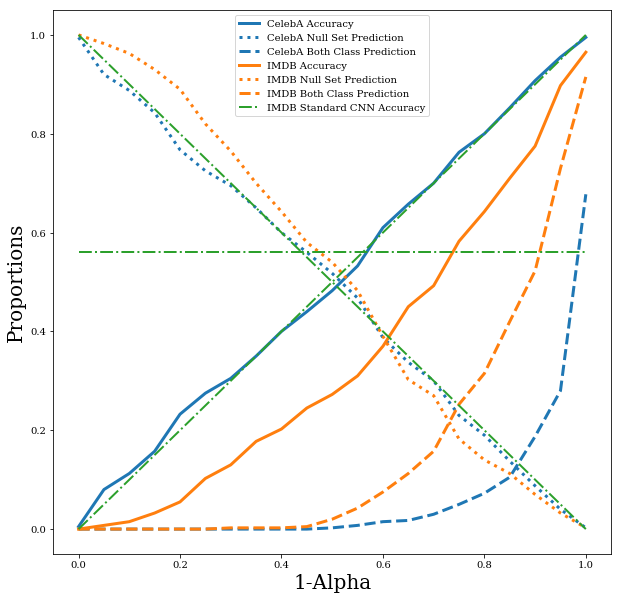} \\
(a) & (b) & (c)\\[6pt]
\end{tabular}
\caption{(a) Females faces mean pixel values in (a) CelebA; (b) IMDB-Wiki. Within CelebA the pictures are aliened with fixed posture, explaining why it naively fails to generalize to IMDB-Wiki images. (c) Performance plots for the conformal method on both CelebA and IMDB-Wiki.}
\label{fig:Faces_Example}
\end{figure*}

We train a standard convolutional neural network ($5$ convolution and $2$ dense layers with the corresponding pooling and activation layers) to perform gender classification on CelebA. It converges well obtaining $0.963$ accuracy on a held out test set, but fails to generalize to the IMDB-Wiki dataset achieving $0.577$ accuracy, slightly better than a random guess. The discrepancy between the two datasets follows from the fact that facial images are reliant on preprocessing to standardize the input. We have used the default preprocessing provided by the datasets, to reflect a scenarios in which the distribution of the samples changes between the training and the testing. Figure \ref{fig:Faces_Example} (a) and (b) show mean pixel values for females pictures within CelebA vs pictures in the IMDB-Wiki dataset. As seen, the IMDB-Wiki is richer and offers larger variety of human postures. 

Although the standard classification method fails in this scenario, the conformal method suggested in this paper still offers valid and sensible results both on CelebA and IMDB-Wiki when using the features extracted from the network trained on CelebA. Figure \ref{fig:Faces_Example} (c) shows the performance of the method with respect to both datasets. CelebA results are good since they are based on features that perform well for this dataset. The level of accuracy is roughly $1-\alpha$ as expected by the design, while the proportion of null predictions is roughly $\alpha$. Therefore for all $\alpha$ there are almost no false positives and all of the errors are the null set. 

The IMDB-Wiki results are not as good, but better than naively using a $0.577$ accuracy classifier. Figure \ref{fig:Faces_Example} (c) show the classifier performance as a function of $\alpha$. Both the accuracy and the number of false positives are tunable. For high values of $1-\alpha$ the accuracy is much higher than $0.577$, but would results in a large number of observations predicted as both genders. If cautious and conservative prediction is required small values of $1-\alpha$ would guarantee smaller number of false predictions, but a large number of null predictions. The suggested conformal method provides a hyper-parameter controlling which type of errors are created according to the prediction needs, and works even in cases where standard methods fail. 

\section{Discussion}\label{sec:Discussion}

In this paper we 
showed that conformal, set-valued predictors
based on
$\hat p(x|y)$
have very good properties.
We obtain a
cautious prediction associating an
observation with a class only if the there is high probability of that
observation is generated from the class. In most of the space the
classifier predicts the null set. This stands in contrast to standard
solutions which provide confident predictions for the entire space
based on data observed from a small area. This can be useful when a
large number of outliers are expected or in which the distribution of
the training data won't fully describe the distribution of the
observations when deployed. Adversarial attacks are an important example
of such scenarios. We also obtain a large set of labels in the set when the object
is ambiguous and is consistent with many different classes.
Thus, our method quantifies two types of uncertainty:
ambiguity with respect to the given classes and outlyingness
with respect to the given classes. 

In addition, the conformal framework provides our method with its
coverage guarantees and class adaptivity. It is straightforward to add
and remove classes at any stage of the process while controlling
either the overall or class specific coverage level of the method in a
highly flexible manner, if desired by the application. This desired properties comes with a
price. The distribution of $p(x|y)$ for each class is learned
independently and the decision boundaries are indifferent to data not
within the class. In case precise decision boundries are more desired, complex estimation functions
overcome this limitation and provide decision boundaries almost equivalent to standard methods.

During the deployment of the method, evaluation of a large number of
kernel density estimators is required. This is relatively slow compared to current methods. This issue can be
addressed in future research with more efficient ways to learn 
ordering-consistent approximations of $p(x|y)$ that can be deployed on GPU's.

\newpage


\newpage

\appendix

\section{Appendix: Details on Proposition 1}
Here we provide more details on Proposition 1.
We assume that the conditions in \cite{cadre2009clustering} hold.
In particular, we assume that
$n h_y^d/(\log n)^{16}\to \infty$ and
$n h_y^{d+4} (\log n)^2 \to 0$
where $h_y$ is the bandwidth of the density estimator.
In addition we assume that ${\cal X}$ is compact
and that $\min_y n_y \to\infty$
where
$n_y = \sum I(Y_i=y)$.

Let
$C(x) = \{y:\ p(x|y) > t_y\}$
and
$\hat C(x) = \{y:\ \hat p(x|y) > t_y\}$.
Note that, conditional on the training data ${\cal D}$,
\begin{align*}
P(Y\in \hat C(X)) &=
\sum_y \int I(y\in \hat C(x)) p(x|y)dx\\
&=
\sum_y \int I(y\in \hat C(x)) \hat p(x|y) dx +
\sum_y \int I(y\in \hat C(x)) [p(x|y)-\hat p(x|y)] dx .
\end{align*}
From Theorem 2.3 of
\cite{cadre2009clustering} 
we have that
$\mu( \{ p(x|y) \geq t\} \Delta \{ p(x|y) \geq \hat t\} ) = O_P(\sqrt{1/(n_y h_y^d)}) = o_P(1)$
where $\mu$ is Lebesgue measure and
$\Delta$ denotes the set difference.
It follows that
$$
\int I(y\in \hat C(x)) p(x|y)dx =\int I(y\in  C(x)) p(x|y)dx + o_P(1) =
1-\alpha + o_P(1).
$$
Also,
\begin{align*}
\Biggl|\sum_y \int I(y\in \hat C(x)) [p(x|y)-\hat p(x|y)] dx \Biggr|  &\leq
\sum_y \int I(y\in \hat C(x)) |p(x|y)-\hat p(x|y)| dx\\
& \leq
k \max_y ||\hat p(x|y) - p(x|y)||_\infty \stackrel{P}{\to} 0
\end{align*}
since, under the conditions,
$\hat p(x|y)$ is consistent in the $\ell_\infty$ norm.
It follows that
$P(Y\in \hat C(X)) = 1-\alpha + o_P(1)$
as required.

We should remark that, in the above,
we assumed that the number of classes is fixed.
If we allow $k$ to grow
the analysis has to change.
Summing the errors in the expression above we have that
$P(Y\in \hat C(X)) = 1-\alpha + R$
where now the remainder is

$$
R = 
O\left(
\sum_y \frac{1}{\sqrt{n_h h_h^d}} +
\sum_y \left(\frac{\log n_y}{n_y}\right)^\frac{2}{4+d}\right).
$$

We then need assume that as $k$ increases,
the $n_y$ grow fast enough so that
$R\stackrel{P}{\to} 0$.
However, this condition can be weakened by
insisting that for all $y$ with $n_y$ small,
we force $\hat C$ to omit $y$.
If this is done carefully, then the coverage condition can be preserved
and we only need 
$R$ to be small when summing over the larger classes.
The details of the theory in this case
will be reported in future work.

\newpage

\end{document}